\documentclass[lettersize,journal]{IEEEtran}

\usepackage{amsmath,amsfonts}
\usepackage{algorithmic}
\usepackage{algorithm}
\usepackage{array}
\usepackage[caption=false,font=normalsize,labelfont=sf,textfont=sf]{subfig}
\usepackage{textcomp}
\usepackage{stfloats}
\usepackage{booktabs}
\usepackage{url}
\usepackage{verbatim}
\usepackage{graphicx}
\usepackage{multirow}
\usepackage[normalem]{ulem}
\usepackage{cite}
\usepackage{balance}

\usepackage[table,dvipsnames]{xcolor}
\definecolor{citecolor}{RGB}{66,168,235}
\definecolor{linkpink}{HTML}{D683B7}

\usepackage{hyperref}
\hypersetup{
    colorlinks=true,
    citecolor=citecolor,
    linkcolor=linkpink,
    urlcolor=linkpink
}

\def\BibTeX{{\rm B\kern-.05em{\sc i\kern-.025em b}\kern-.08em
    T\kern-.1667em\lower.7ex\hbox{E}\kern-.125emX}}
\usepackage{balance}
\begin{document}

\title{RSJEV: Discriminative Remote Sensing Scene Classification with Multimodal Large Language Models}

\author{Dongchen~Si,
        Di~Wang,
        Mingzhen~Xu,
        Jing~Zhang,~\IEEEmembership{Senior~Member,~IEEE},
        Bo~Du,~\IEEEmembership{Senior~Member,~IEEE},
        and Liangpei Zhang,~\IEEEmembership{Fellow,~IEEE}
        
\thanks{\textit{Corresponding authors: Di Wang and Bo~Du}}

\thanks{Dongchen Si, Di Wang, Mingzhen Xu, Jing Zhang and Bo Du are with the School of Computer Science, Wuhan University, Wuhan 430072, China (e-mail: dochsi@outlook.com; d\_wang@whu.edu.cn; mingzhenxu@whu.edu.cn; jingzhang.cv@whu.edu.cn; dubo@whu.edu.cn).}
\thanks{Liangpei Zhang is with the State Key Laboratory of Information Engineering in Surveying, Mapping and Remote Sensing, Wuhan University, Wuhan 430079, China (e-mail: zlp62@whu.edu.cn).}
}



\maketitle
\begin{abstract}

Remote sensing scene classification is a fundamental task in Earth observation and geospatial analysis. Existing approaches mainly follow three paradigms: task-specific visual classification, vision-language similarity matching, and autoregressive multimodal generation. However, visual classifiers rely on predefined label spaces, CLIP-based methods perform recognition through static image-text alignment, and multimodal large language models (MLLMs) introduce unnecessary token-level generation for classification tasks with explicit candidate categories. To address these limitations, we propose RSJEV, a one-pass multimodal decision framework for remote sensing scene classification. Unlike conventional MLLMs that formulate classification as autoregressive text generation, RSJEV reformulates scene classification as a candidate-conditioned multimodal discriminative decision process, where visual representations, task instructions, and candidate category semantics are jointly modeled. Specifically, we introduce a OnePass Decider that extracts multimodal decision states and directly estimates category probabilities within the candidate category space, eliminating autoregressive decoding while preserving vision-language interactions. Extensive experiments on three widely used remote sensing scene classification benchmarks, including UC Merced, AID, and NWPU-RESISC45, demonstrate that RSJEV achieves superior classification performance compared with representative CNN-, Transformer-, Mamba-, CLIP-, and MLLM-based methods. Moreover, RSJEV significantly reduces inference costs and achieves a better accuracy-efficiency trade-off with only a compact 0.8B-parameter model. These results demonstrate the effectiveness of state-conditioned multimodal decision making for efficient remote sensing image understanding. The code will be available at \href{https://github.com/Dongtcs/RSJEV}{\textcolor{linkpink}{https://github.com/Dongtcs/RSJEV}}.

\end{abstract}

\IEEEpeerreviewmaketitle
\section{Introduction}

\IEEEPARstart{R}emote Sensing (RS) is an essential technology for Earth observation \cite{chintalapati2025opportunities}, environmental monitoring \cite{shu2025deep}, disaster assessment \cite{al2024integrating}, and geospatial intelligence \cite{liu2024explainable}. As a fundamental task in RS image interpretation, scene classification provides scene-level semantic information to support various geospatial applications. However, substantial intra-class variability, inter-class visual similarity, and differences in spatial resolution pose challenges to accurate RS scene classification.

In recent years, convolutional neural networks (CNNs) \cite{lecun1998gradient} and vision transformers (ViTs) \cite{dosovitskiy2020vit} have substantially improved RS scene classification \cite{He_2016_CVPR,huang2017densely,RSVA,rsp}. CNNs capture local spatial patterns through hierarchical feature extraction, whereas ViTs model long-range dependencies through self-attention. These models typically use task-specific classification heads that map visual representations to a predefined label space. Although effective for supervised classification, this formulation does not explicitly incorporate candidate category semantics expressed in natural language into the prediction process. Consequently, it lacks rich semantic priors to disambiguate visually confusing categories—such as differentiating visually subtle land covers (e.g., \textit{desert} versus \textit{barren land}) or geometrically akin infrastructures (e.g., \textit{railway} versus \textit{freeway})—and strictly restricts the model to a fixed, non-transferable label space.

Vision-language models (VLMs) \cite{radford2021learning,khattakMaPLe,11192766} introduce textual category semantics into RS scene classification. CLIP \cite{radford2021learning} aligns images and natural language in a shared embedding space, enabling recognition through textual descriptions of candidate categories. CLIP-based approaches, including prompt-learning variants \cite{khattakMaPLe,11192766}, typically compute category scores using similarities between image and text embeddings. This formulation incorporates category semantics through embedding similarity, but provides limited direct interaction among visual features, task instructions, and candidate categories during prediction. Moreover, classification performance can depend on how category prompts are formulated.

Multimodal large language models (MLLMs) \cite{bai2025qwen3vltechnicalreport, zhu2025internvl3exploringadvancedtraining,LI2025539,kuckreja2023geochat} integrate visual representations with language modeling to support instruction-guided image interpretation. When applied to RS scene classification, generative MLLMs typically predict categories by producing textual responses autoregressively. Although this formulation supports flexible outputs, generating category names or explanatory answers can require multiple decoding steps, increasing inference latency and computational cost. For classification tasks with an explicit candidate category set, the required output is a discrete category decision, motivating prediction mechanisms that avoid iterative text generation.

\begin{figure*}[t]
    \centering
    \includegraphics[width=\textwidth]{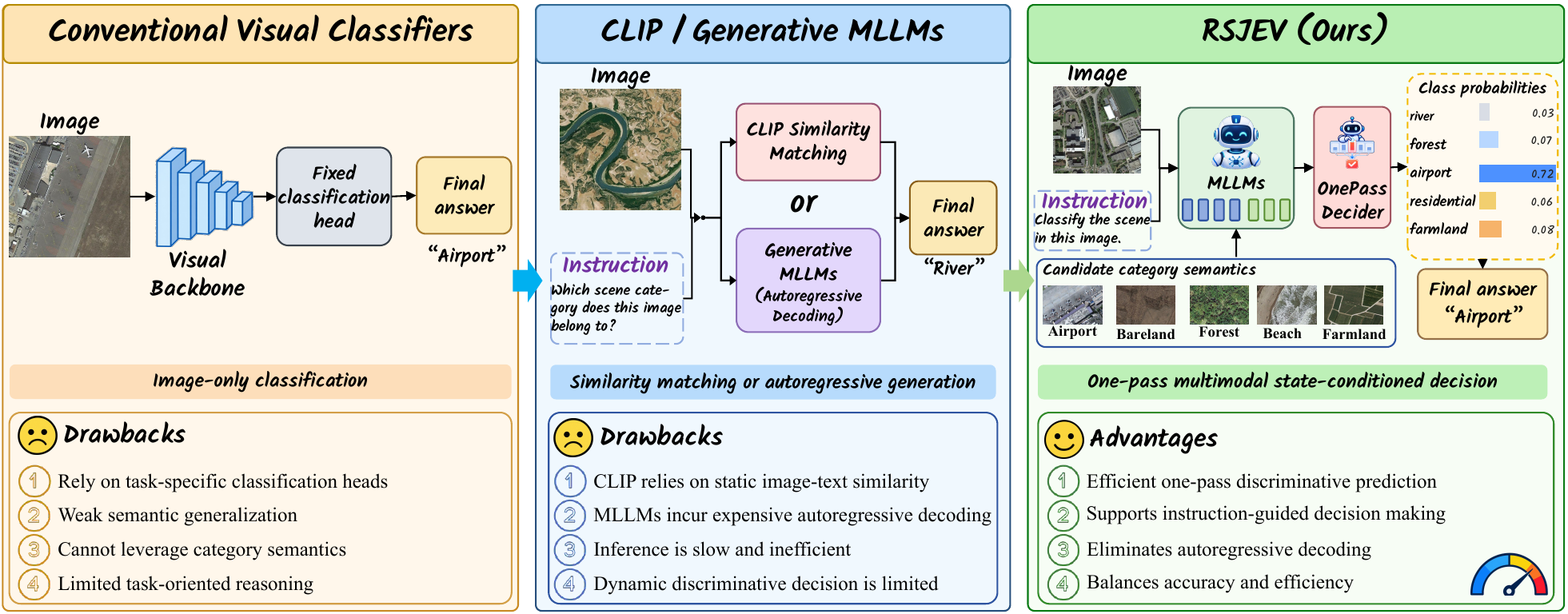}
    \caption{Conceptual comparison of different classification paradigms: (left) conventional image-only visual classifiers, (middle) similarity-based or generative MLLMs, and (right) the proposed RSJEV. By eliminating iterative autoregressive decoding, RSJEV enables instruction-guided one-pass discriminative decision-making, achieving an effective balance between classification performance and inference efficiency.}
    \label{fig:introduction}
\end{figure*}

To retain multimodal interactions while avoiding iterative text generation, we propose RSJEV, a one-pass multimodal decision framework for RS scene classification (Fig.~\ref{fig:introduction}). RSJEV reformulates classification as a candidate-conditioned discriminative decision problem, jointly modeling visual representations, task instructions, and candidate category semantics. Specifically, we introduce OnePass Decider, which extracts the final-layer hidden state at a designated answer slot as a multimodal decision representation. It then reuses the pretrained language modeling head to score candidate options and computes category probabilities through a softmax over the candidate space. This design enables direct category prediction in a single forward pass without autoregressive answer generation.

\vspace{4pt}
The primary contributions of this paper are summarized as follows:
\begin{itemize}
    \item We propose RSJEV, a one-pass multimodal decision framework for RS scene classification. It reformulates classification as a candidate-conditioned discriminative decision problem, enabling direct category prediction without autoregressive answer generation.
    
    \item We introduce OnePass Decider, which extracts an answer-slot decision representation conditioned on visual content, task instructions, and candidate category semantics. It reuses the pretrained language modeling head to score candidate options and estimate probabilities within the specified candidate space.
    
    \item Extensive experiments on three benchmark datasets alongside comprehensive ablations demonstrate that RSJEV achieves competitive accuracy with significantly lower inference latency compared to generative MLLMs.
\end{itemize}

\section{Related Work}

Remote sensing image classification has been extensively studied with the development of deep learning. CNN-based models, such as ResNet \cite{He_2016_CVPR}, DenseNet \cite{huang2017densely}, and EfficientNet \cite{pmlr-v97-tan19a}, have achieved remarkable performance by learning hierarchical visual representations. Recently, Transformer-based architectures, including Vision Transformer \cite{dosovitskiy2020vit} and Swin Transformer \cite{liu2021swin}, have further improved scene classification by modeling global contextual relationships. Meanwhile, emerging state-space models, such as RSMamba \cite{RSMamba}, have explored efficient global feature modeling for remote sensing images. Despite their effectiveness, these methods generally rely on task-specific classification heads and predefined category spaces, which limits their adaptability to dynamic and open-vocabulary scenarios.

Contrastive vision-language models incorporate textual semantics into remote sensing image recognition. CLIP \cite{radford2021learning} aligns image and text representations in a shared embedding space, enabling classification using natural-language descriptions of candidate categories. Building on this framework, RemoteCLIP \cite{liu2024remoteclip} expands remote sensing pretraining data by converting detection and segmentation annotations into image-caption pairs. SkyScript \cite{wang2024skyscript} constructs remote sensing image-text pairs using semantic information from OpenStreetMap and supports continual pretraining of SkyCLIP. These efforts improve image-text alignment for the remote sensing domain. In their standard zero-shot classification setting, category predictions are obtained by comparing image embeddings with text embeddings of candidate categories. This similarity-based formulation differs from constructing a joint decision representation conditioned on image content, task instructions, and the candidate set.

Multimodal large language models combine visual encoders with language models to support image-conditioned language tasks. General models, including BLIP-2 \cite{li2023blip}, LLaVA \cite{liu2023llava}, Qwen3-VL \cite{bai2025qwen3vltechnicalreport}, and InternVL3 \cite{zhu2025internvl3exploringadvancedtraining}, provide a foundation for multimodal image interpretation. In remote sensing, GeoChat \cite{kuckreja2023geochat} supports image- and region-level conversations and visual grounding. LHRS-Bot \cite{muhtar2024lhrs} leverages volunteered geographic information to construct training data and introduces a multi-level vision-language alignment strategy. EarthGPT \cite{zhang2024earthgpt} integrates interpretation tasks across optical, synthetic aperture radar, and infrared imagery through multisensor instruction tuning. These models primarily formulate task outputs as autoregressively generated responses. For scene classification with explicit candidate categories, generating category names or explanatory answers can require additional decoding steps. RSJEV instead extracts a candidate-conditioned decision representation at an answer slot and directly scores candidate options, enabling classification without iterative answer generation.

The reviewed methods primarily focus on visual representation learning, contrastive image-text alignment, or generative multimodal interpretation. RSJEV adopts a candidate-conditioned discriminative formulation that extracts an answer-slot representation from the multimodal backbone and reuses the pretrained language modeling head to score candidate options. By conditioning the decision representation on image content, task instructions, and candidate category semantics, RSJEV directly estimates category probabilities in a single forward pass without iterative text generation.
\section{Methodology}
\subsection{Problem Formulation}

Given a remote sensing image $I_i$ and an ordered list of candidate categories $\mathcal{C}=(c_1,c_2,\ldots,c_K)$, we formulate scene classification as a candidate-conditioned multimodal discriminative decision problem. Each labeled sample is represented as
\begin{equation}
X_i=(I_i,P_i,t_i),
\end{equation}
where $P_i$ is the task prompt and $t_i\in\{1,\ldots,K\}$ denotes the index of the ground-truth category in $\mathcal{C}$. The prompt comprises a classification question and its candidate options:
\begin{equation}
P_i=[\text{Question},\text{Options}].
\end{equation}

The candidate category descriptions are included in the prompt as semantic conditions and processed jointly with the image by the multimodal backbone. Rather than generating a category name through autoregressive decoding, RSJEV directly estimates a probability distribution over the $K$ candidate options:
\begin{equation}
p_{i,k}=p_{\theta}(k\mid I_i,P_i,\mathcal{C}),
\qquad k=1,\ldots,K,
\end{equation}
where $\theta$ denotes the model parameters and $\sum_{k=1}^{K}p_{i,k}=1$. The predicted category is obtained by selecting the option with the highest probability:
\begin{equation}
\hat{t}_i=\arg\max_{k\in\{1,\ldots,K\}}p_{i,k},
\qquad
\hat{y}_i=c_{\hat{t}_i}.
\end{equation}

This formulation enables direct classification within the supplied candidate category space through a single forward pass, without autoregressive answer generation.

\subsection{RSJEV}
In the original JEV \cite{typesafe_api}, the same question needs to be expanded into multiple input paths according to the number of candidate options, and each path is separately fed into the model. For remote sensing image classification, this process requires repeated computation of image features and textual task prompts, introducing additional computational overhead. To address this issue, we propose RSJEV, which completes task decision through a single forward pass with one input path. The details are introduced as follows:
\subsubsection{Overall Workflow} 
\begin{figure*}[t]
\centering
\resizebox{0.99\linewidth}{!}{
\includegraphics[width=\linewidth]{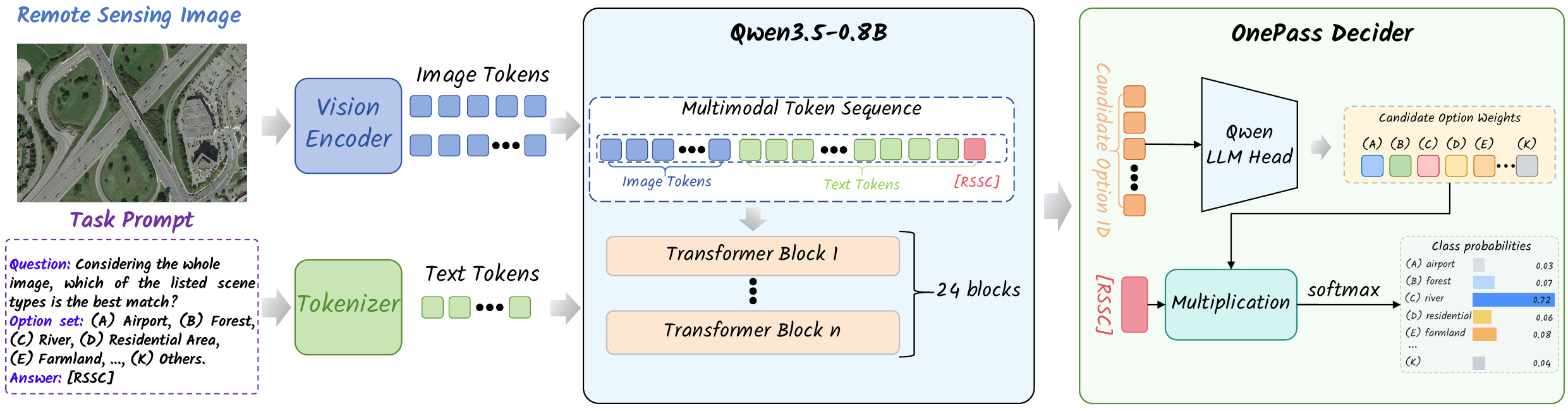}
}
\vspace{-6pt}
\caption{Overview of the proposed RSJEV framework for one-pass remote sensing scene classification.
}\label{fig:RSJEV}
\vspace{-10pt}
\end{figure*}

As illustrated in Fig.~\ref{fig:RSJEV}, RSJEV combines a pretrained multimodal backbone with OnePass Decider (OPD) for one-pass scene classification. The backbone jointly processes the image and a task prompt containing the instruction and candidate categories. OPD extracts the final-layer hidden state at the classification answer slot and reuses the pretrained language modeling head to compute candidate-option scores. Softmax converts these scores into category probabilities, and the highest-scoring option determines the prediction without autoregressive answer generation.

\subsubsection{Model Components}

\textbf{MLLM Backbone:} We adopt pretrained multimodal large language models as the multimodal backbone of RSJEV. The backbone jointly processes the remote sensing image and the task prompt containing the classification instruction and candidate category descriptions, producing contextualized multimodal hidden states. OPD then uses the final-layer hidden state at the classification answer slot for discriminative prediction.

\textbf{OnePass Decider:} To exploit multimodal contextual representations for direct classification while avoiding autoregressive decoding overhead, we introduce OPD for candidate-conditioned discriminative decision-making. OPD uses the final-layer hidden state at the designated \texttt{[RSSC]} answer slot to directly estimate candidate category probabilities within a single forward pass, without iterative answer-token generation. Let the final-layer hidden states of the multimodal backbone be
\begin{equation}
H_i=[h_{i,1},h_{i,2},\ldots,h_{i,N_i}]^{\top}
\in\mathbb{R}^{N_i\times d},
\end{equation}
where $N_i$ is the input sequence length and $d$ is the hidden dimension. OPD extracts the decision representation at the designated \texttt{[RSSC]} classification answer slot:
\begin{equation}
h_i^{\mathrm{ans}}=H_i[s_i,:]^{\top} \in\mathbb{R}^{d\times 1},
\end{equation}
where $s_i$ denotes the position of \texttt{[RSSC]} in the input sequence. This representation is conditioned on the image, task instruction, and candidate category descriptions.

Each candidate option is associated with a token id $u_k$. Then, OPD selects the corresponding weights from the pretrained language modeling head:
\begin{equation}
W_{\mathcal{C}}=
\begin{bmatrix}
W_{\mathrm{LM}}[u_1,:]\\
\vdots\\
W_{\mathrm{LM}}[u_K,:]
\end{bmatrix},
\qquad
z_i=W_{\mathcal{C}}h_i^{\mathrm{ans}}.
\end{equation}
The candidate probabilities are obtained by normalizing only the selected scores:
\begin{equation}
p_{i,k}=
\frac{\exp(z_{i,k})}
{\sum_{j=1}^{K}\exp(z_{i,j})}.
\end{equation}
where $W_{\mathrm{LM}} \in \mathbb{R}^{L \times d}$ denotes the weight matrix of the pretrained language modeling head, and $L$ is the vocabulary size.

The highest-probability option determines the predicted category, completing classification within a single forward pass without autoregressive answer generation.

\section{Experimental Results and Analysis}

In this section, we evaluate RSJEV on three remote sensing scene classification benchmarks: UCM, AID, and NWPU-RESISC45. After describing the datasets and implementation details, we compare its classification performance with representative baselines. We then assess the applicability of OPD across different MLLM backbones. Finally, we compare RSJEV with generative MLLM baselines in terms of inference latency, GPU memory consumption, and throughput.

\subsection{Dataset Description}
We evaluate RSJEV on three widely used remote sensing scene classification datasets: UC Merced Land Use (UCM) \cite{yang2010bag}, Aerial Image Dataset (AID) \cite{xia2017aid}, and NWPU-RESISC45 (NWPU) \cite{cheng2017remote}. Following the split protocol in \cite{RSVA}, we adopt the UCM-55, AID-28, and NWPU-28 settings, which use 50\%, 20\%, and 20\% of the images for training, respectively, with the remaining images used for testing.

\textbf{UCM} \cite{yang2010bag}: 
The dataset contains 2,100 aerial images spanning 21 scene categories, with 100 images per category. The images have a spatial resolution of approximately 0.3 m per pixel and are typically $256 \times 256$ pixels in size.

\textbf{AID} \cite{xia2017aid}: 
The dataset comprises 10,000 aerial images collected from Google Earth, covering 30 scene categories with 220–420 images per category. Each image is $600 \times 600$ pixels in size, with spatial resolutions ranging from 0.5 to 8 m per pixel.

\textbf{NWPU} \cite{cheng2017remote}: 
The dataset contains 31,500 RGB remote sensing images collected from Google Earth, spanning 45 scene categories with 700 images per category. Each image is $256 \times 256$ pixels in size.

\subsection{Implementation Details}

\begin{table*}[!tbhp]
\centering
\caption{Comparison with state-of-the-art methods on three remote sensing scene classification benchmarks. The \textbf{bold} and \underline{underlined} values denote the best and second-best results, respectively.}
\label{tab:rsjev_compare}

\resizebox{0.9\linewidth}{!}{
\begin{tabular}{*2{l}| *4{c}| *4{c}| *4{c}}
\toprule
\multirow{2}{*}{Method} & \multirow{2}{*}{Param} 
& \multicolumn{4}{c|}{UCM} 
& \multicolumn{4}{c|}{AID} 
& \multicolumn{4}{c}{NWPU} \\

& & P & R & F1 & OA 
& P & R & F1 & OA 
& P & R & F1 & OA \\

\midrule  

\multicolumn{14}{l}{\textcolor{gray}{\textit{CNN-based Models}}} \\
ResNet-101 \cite{He_2016_CVPR} & 42.6 M 
& 95.91 & 95.71 & 95.70 & 95.71
& 92.33 & 91.79 & 91.89 & 92.23
& 90.34 & 90.34 & 90.25 & 90.34 \\

DenseNet-161 \cite{huang2017densely} & 28.7 M 
& 97.18 & 97.05 & 97.02 & 97.05
& 93.90 & 93.48 & 93.57 & 93.90
& 92.77 & 92.73 & 92.69 & 92.73 \\

EfficientNet \cite{pmlr-v97-tan19a} & 87.4 M 
& 97.12 & 96.76 & 96.80 & 96.76
& 91.48 & 91.39 & 91.33 & 91.76
& 92.49 & 92.45 & 92.43 & 92.45 \\

\midrule
\multicolumn{14}{l}{\textcolor{gray}{\textit{Transformer-based Models}}} \\
DeiT III \cite{10.1007/978-3-031-20053-3_30} & 86.9 M 
& 94.46 & 94.28 & 94.19 & 94.28
& 88.78 & 88.67 & 88.51 & 89.10
& 87.43 & 87.42 & 87.23 & 87.42 \\

Swin Transformer \cite{liu2021swin} & 86.8 M 
& 96.70 & 96.48 & 96.41 & 96.47
& 94.58 & 94.50 & 94.51 & 94.76
& 92.76 & 92.74 & 92.69 & 92.74 \\

Vision Transformer \cite{dosovitskiy2020vit} &  88.3 M 
& 86.48 & 85.81 & 85.87 & 85.81
& 81.50 & 80.53 & 80.69 & 81.19
& 78.14 & 77.93 & 77.79 & 77.93 \\

\midrule
\multicolumn{14}{l}{\textcolor{gray}{\textit{Mamba-based Models}}} \\
VMamba \cite{NEURIPS2024_baa2da9a} & 89.0 M 
& \underline{97.39} & 97.14 & 97.14 & 97.14
& \underline{95.67} & 95.43 & 95.50 & \underline{95.79}
& \underline{94.22} & \underline{94.20} & \underline{94.17} & \underline{94.20} \\

Vision Mamba \cite{vim} & 98.0 M 
& 96.65 & 96.38 & 96.40 & 96.38
& 94.23 & 94.03 & 94.06 & 94.39
& 93.92 & 93.88 & 93.86 & 93.88 \\

RSMamba \cite{RSMamba} &  33.1 M 
& 94.78 & 94.57 & 94.59 & 94.48
& 82.28 & 81.63 & 81.70 & 82.37
& 80.98 & 80.99 & 80.87 & 80.99 \\

\midrule
\multicolumn{14}{l}{\textcolor{gray}{\textit{Vision-Language Models}}} \\
Open-CLIP \cite{radford2021learning} & 151.0 M 
& 96.57 & 96.38 & 96.40 & 96.38
& 95.35 & 95.21 & 95.23 & 95.46
& 91.14 & 90.83 & 90.70 & 91.22 \\

MaPLe \cite{khattakMaPLe} & 151.0 M 
& 95.85 & 95.71 & 95.74 & 95.71
& 92.98 & 92.43 & 92.61 & 92.93
& 88.75 & 88.68 & 88.66 & 88.68 \\

OSCLIP \cite{11192766} & 428.0 M 
& \underline{97.39} & \underline{97.33} & \underline{97.33} & \underline{97.33}
& 95.54 & 95.40 & 95.45 & 95.70
& 93.36 & 93.31 & 93.30 & 93.31 \\

\midrule
\multicolumn{14}{l}{\textcolor{gray}{\textit{Multimodal Large Language Models}}} \\
Qwen3-VL \cite{bai2025qwen3vltechnicalreport} & 2.0 B 
& 96.04 & 95.81 & 95.79 & 95.81
& 95.61 & \underline{95.60} & \underline{95.52} & 95.70
& 92.94 & 92.83 & 92.81 & 92.88 \\

InternVL3 \cite{zhu2025internvl3exploringadvancedtraining} & 2.0 B 
& 94.70 & 94.48 & 94.47 & 94.48
& 94.76 & 94.45 & 94.55 & 94.93
& 92.24 & 92.11 & 92.13 & 92.11 \\

LHRS-Bot-Nova \cite{LI2025539} & 8.0 B 
& 83.90 & 80.48 & 78.99 & 80.47
& 82.48 & 81.04 & 79.71 & 82.36
& 83.79 & 81.01 & 79.30 & 81.01 \\

GeoChat \cite{kuckreja2023geochat} & 7.0 B 
& 86.17 & 78.10 & 77.21 & 78.10
& 79.14 & 73.98 & 71.47 & 73.80
& 91.26 & 90.23 & 90.23 & 90.23 \\

\midrule
RSJEV (Ours) & 0.8 B 
& \textbf{97.80} & \textbf{97.71} & \textbf{97.72} & \textbf{97.71}
& \textbf{96.44} & \textbf{96.22} & \textbf{96.30} & \textbf{96.56}
& \textbf{94.61} & \textbf{94.59} & \textbf{94.58} & \textbf{94.59} \\

\bottomrule
\end{tabular}
}
\end{table*}
\textbf{Experimental Settings:}
The experiments are conducted based on the Qwen3.5-0.8B \cite{bai2025qwen3vltechnicalreport} multimodal model. During training, the model is optimized using the AdamW optimizer with a learning rate of \(1\times10^{-4}\), a weight decay of 0.01, and a learning rate warm-up strategy with 50 warm-up steps. Following existing JEV-style implementations\footnote{https://github.com/Mapika/decider}, we use cross-entropy loss together with Brier score regularization during training. The model is trained for 10 epochs on eight NVIDIA RTX A40 GPUs with a per-GPU batch size of 16, resulting in a total batch size of 128. The input image resolution is dynamically adjusted by the multimodal backbone, with the maximum token length is limited to 1536. Random image flipping is applied during training to improve the generalization ability of the model. The label smoothing coefficient is set to 0.05. All experiments are conducted using FP32 precision.

\textbf{Evaluation Metrics:} We use precision (P), recall (R), F1-score (F1), and overall accuracy (OA) as evaluation metrics.

\subsection{Comparison Experiments}

We compare RSJEV with representative approaches from five categories: CNN-based Classification Models, Transformer-based Classification Models, Mamba-based Classification Models, Vision-Language Models, and Multimodal Large Language Models. Specifically, the CNN-based baselines include ResNet-101 \cite{He_2016_CVPR}, DenseNet-161 \cite{huang2017densely}, and EfficientNet \cite{pmlr-v97-tan19a}. The Transformer-based baselines include DeiT III \cite{10.1007/978-3-031-20053-3_30}, Swin Transformer \cite{liu2021swin}, and Vision Transformer \cite{dosovitskiy2020vit}. The Mamba-based baselines include VMamba \cite{NEURIPS2024_baa2da9a}, Vision Mamba \cite{vim}, and RSMamba \cite{RSMamba}. The Vision-Language Models include Open-CLIP \cite{radford2021learning}, MaPLe \cite{khattakMaPLe}, and OSCLIP \cite{11192766}. The Multimodal Large Language Models include Qwen3-VL \cite{bai2025qwen3vltechnicalreport}, InternVL3 \cite{zhu2025internvl3exploringadvancedtraining}, LHRS-Bot-Nova \cite{LI2025539}, and GeoChat \cite{kuckreja2023geochat}. For CNN-, Transformer-, and Mamba-based models, we initialize the networks with ImageNet-1K pretrained weights provided by MMPreTrain \cite{2023mmpretrain}. For Vision-Language Models and Multimodal Large Language Models, we use their officially released pretrained checkpoints and further fine-tune them on the remote sensing scene classification task.
\subsubsection{Quantitative Analyses}
As reported in Table~\ref{tab:rsjev_compare}, RSJEV achieves the highest precision, recall, F1-score, and overall accuracy across all three benchmarks. Specifically, its F1-scores reach 97.72\%, 96.30\%, and 94.58\% on UCM, AID, and NWPU, respectively. Compared with the best-performing existing method on each dataset: namely OSCLIP on UCM, Qwen3-VL on AID, and VMamba on NWPU, RSJEV achieves improvements of 0.39, 0.78, and 0.41 percentage points, respectively. Furthermore, RSJEV consistently outperforms all evaluated contrastive vision-language models and generative MLLMs. Crucially, these gains are attained with a compact 0.8B-parameter backbone, outperforming scaled generative counterparts spanning 2B to 8B parameters. These empirical comparisons firmly demonstrate that RSJEV delivers superior classification performance while maintaining remarkable parameter efficiency.

\subsection{Generalizability of OnePass Decider}

We evaluate the effectiveness and generalizability of OPD across different MLLM backbones, with detailed results reported in Table~\ref{tab:mllm_generalization}. Baseline denotes the generative classification configuration of each backbone, i.e., standard autoregressive next-token prediction of category answers, whereas Ours denotes the corresponding model equipped with OPD. On Qwen3.5-0.8B, OPD improves precision from 87.96\%, 91.29\%, and 88.24\% to 97.80\%, 96.44\%, and 94.61\% on UCM, AID, and NWPU, respectively, corresponding to gains of 9.84, 5.15, and 6.37 percentage points. To further assess whether these improvements generalize beyond the backbone used in the main experiments, we additionally adapt OPD to InternVL3.5-1B. OPD again consistently improves precision across all three benchmarks, yielding gains of 3.47, 4.01, and 1.39 percentage points on UCM, AID, and NWPU, respectively. These consistent improvements across both Qwen3.5-0.8B and InternVL3.5-1B demonstrate that OPD is not tied to a specific MLLM backbone and exhibits strong generalizability across different architectures.

\subsection{Inference Efficiency Analysis}

\begin{figure*}[h]
    \centering
    \begin{minipage}[t]{0.49\textwidth}
        \centering
        \includegraphics[width=\linewidth]{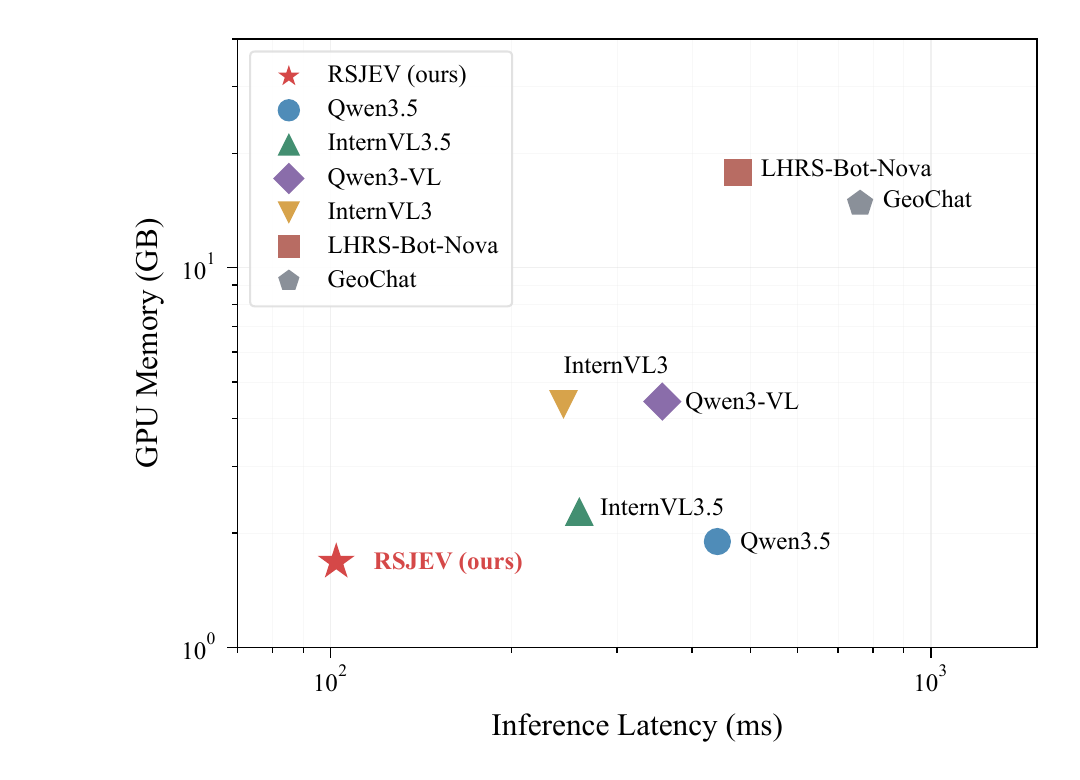}
        \caption{Trade-off between GPU memory consumption and inference latency among different MLLM-based approaches.}
        \label{fig:eff_memory_latency}
    \end{minipage}
    \hfill
    \begin{minipage}[t]{0.49\textwidth}
        \centering
        \includegraphics[width=\linewidth]{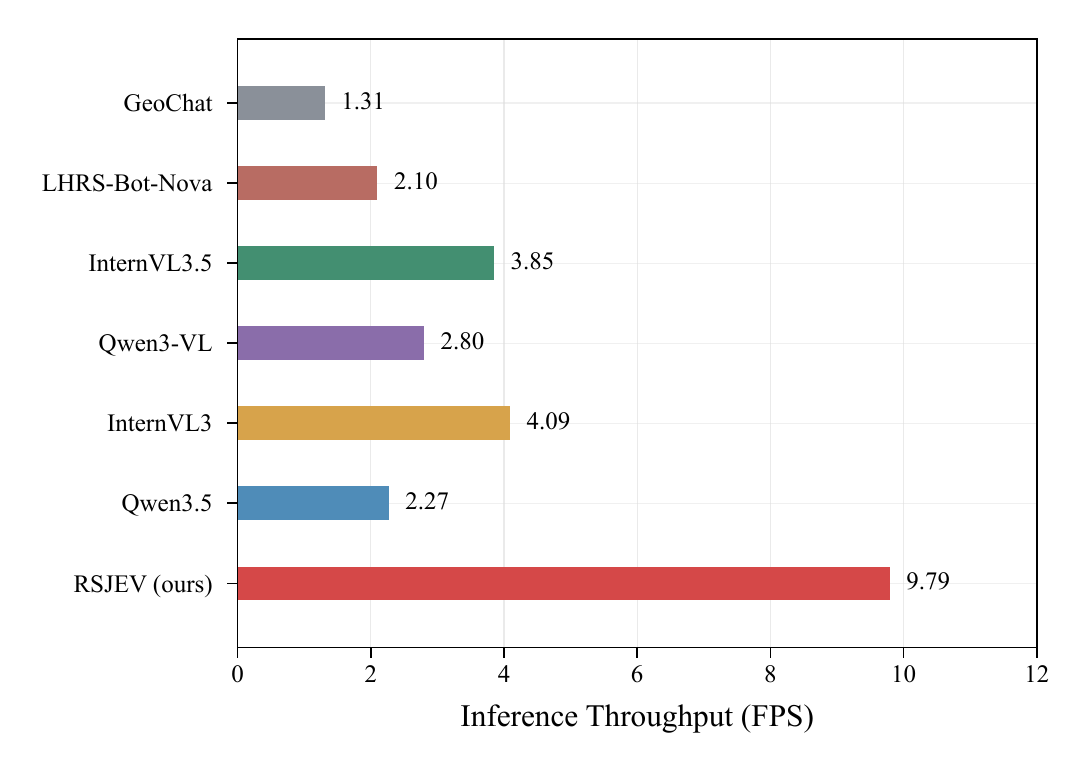}
        \caption{Inference throughput comparison among different MLLM-based approaches in terms of FPS.}
        \label{fig:eff_fps}
    \end{minipage}
\end{figure*}

We evaluate the inference efficiency of RSJEV against six MLLM baselines: Qwen3.5-0.8B, Qwen3-VL-2B, InternVL3.5-1B, InternVL3-2B, LHRS-Bot-Nova-8B, and GeoChat-7B. All measurements are conducted on a single NVIDIA A40 GPU in bfloat16 precision with a batch size of 1, using 30 test images after a five-image warm-up. GPU memory consumption and inference latency are reported in Fig.~\ref{fig:eff_memory_latency}, while throughput is presented in Fig.~\ref{fig:eff_fps}. RSJEV achieves the lowest GPU memory consumption and inference latency among the evaluated methods, requiring 1.68 GB of GPU memory and 102.18 ms per image. Its throughput reaches 9.79 FPS, exceeding all generative baselines, whose throughput ranges from 1.31 to 4.09 FPS. Specifically, RSJEV achieves 4.31 times the throughput of Qwen3.5 and 2.39 times that of InternVL3, the fastest generative baseline. These results highlight the efficiency of the compact RSJEV model, which directly estimates candidate category probabilities through a single forward pass without autoregressive answer generation. Together with the classification results in Table~\ref{tab:rsjev_compare}, these findings demonstrate that RSJEV combines strong classification performance with lower memory requirements and faster inference than the evaluated MLLM baselines, offering a favorable performance-efficiency balance for remote sensing scene classification.

\begin{table}[t]
\centering
\caption{Generalizability across MLLM Backbones (P, \%).}
\label{tab:mllm_generalization}
\small
\setlength{\tabcolsep}{0pt}
\begin{tabular*}{\columnwidth}{@{\extracolsep{\fill}}ccccc}
\toprule
Backbone & Method & UCM & AID & NWPU \\

\midrule

\multirow{2}{*}{Qwen3.5-0.8B}
& Baseline & 87.96 & 91.29 & 88.24 \\
& Ours     & 97.80 & 96.44 & 94.61 \\

\midrule

\multirow{2}{*}{InternVL3.5-1B}
& Baseline & 93.21 & 92.26 & 93.54 \\
& Ours     & 96.68 & 96.27 & 94.93 \\

\bottomrule
\end{tabular*}
\end{table}

\section{Conclusion}

In this paper, we propose RSJEV, a one-pass multimodal decision framework for remote sensing scene classification. RSJEV reformulates classification as a candidate-conditioned multimodal discriminative decision problem by jointly modeling visual representations, task instructions, and candidate category semantics. The proposed OnePass Decider directly estimates candidate category probabilities from the final-layer hidden state at the classification answer slot, avoiding the additional decoding overhead of autoregressive answer generation. Experiments demonstrate that RSJEV achieves strong classification performance while substantially improving inference efficiency over generative MLLM baselines. Cross-backbone experiments further validate the effectiveness and generalizability of OnePass Decider across different MLLM architectures. Consequently, RSJEV highlights the considerable potential of JEV-style multimodal discriminative modeling as a new decision paradigm for remote sensing, offering a promising direction toward more efficient and scalable multimodal foundation models.
\ifCLASSOPTIONcaptionsoff
  \newpage
\fi

\bibliographystyle{IEEEtran}
\normalem
\bibliography{references.bib}
\end{document}